\documentclass[]{utexas}

\usepackage[utf8]{inputenc} 
\usepackage{url}            
\usepackage{amsfonts}       
\usepackage{nicefrac}       

\usepackage{amsmath}
\usepackage{enumerate}
\usepackage{algorithm}
\usepackage[noend]{algpseudocode}
\usepackage{amsthm}
\usepackage{newtxtt}
\usepackage{diagbox}
\usepackage{colortbl}

\usepackage[framemethod=tikz]{mdframed}
\usepackage{amssymb}
\usepackage{xspace}
\usepackage{wrapfig}
\usepackage{adjustbox}
\usepackage{tabularx}
\usepackage{mathtools}

\usepackage{array}
\usepackage[normalem]{ulem} 
\usepackage{makecell}
\usepackage{enumitem}
\usepackage{float}
\usepackage{comment}

\usepackage{tikz}
\usetikzlibrary{arrows.meta,positioning,fit,calc,shapes.misc}
\usetikzlibrary{decorations.markings, backgrounds, shapes.geometric}

\definecolor{lumaBlue}{RGB}{44,115,210}    
\definecolor{lumaGreen}{RGB}{48,175,123}   
\definecolor{lumaRed}{RGB}{225,85,75}      
\definecolor{lumaGray}{RGB}{75,75,75}      
\definecolor{lightGray}{gray}{0.96}        
\definecolor{deepGray}{RGB}{50,50,50}      
\definecolor{lumaOrange}{RGB}{230,150,50}  

\definecolor{FrozenBlue}{RGB}{196,219,255}
\definecolor{TrainPeach}{RGB}{255,229,204}
\definecolor{EmbedGreen}{RGB}{197,224,180}
\definecolor{Ink}{RGB}{30,42,70}

\definecolor{OursGreen}{RGB}{232,245,233}
\definecolor{ColGray}{RGB}{245,245,245}
\definecolor{HeaderGray}{RGB}{250,250,250}

\definecolor{ablRow1}{RGB}{236,245,255}
\definecolor{ablRow2}{RGB}{221,238,255}
\definecolor{ablRow3}{RGB}{204,231,255}
\definecolor{ablRow4}{RGB}{187,224,255}
\definecolor{ablRow5}{RGB}{168,216,255}
\definecolor{ablRow8}{RGB}{130,200,255}

\tcbuselibrary{skins, breakable}
\tcbset{
  lumaBox/.style={
    enhanced,
    colback=lumaBlue!5!white,
    colframe=lumaBlue!60!black,
    fonttitle=\bfseries,
    coltitle=black,
    boxrule=0.5pt,
    arc=2mm,
    left=4pt,
    right=4pt,
    top=3pt,
    bottom=3pt,
    breakable,
  },
  lumaBoxGreen/.style={
    enhanced,
    colback=lumaGreen!5!white,
    colframe=lumaGreen!60!black,
    fonttitle=\bfseries,
    boxrule=0.5pt,
    arc=2mm,
    left=4pt,
    right=4pt,
    top=3pt,
    bottom=3pt,
    breakable,
  },
  lumaBoxRed/.style={
    enhanced,
    colback=lumaRed!5!white,
    colframe=lumaRed!60!black,
    fonttitle=\bfseries,
    boxrule=0.5pt,
    arc=2mm,
    left=4pt,
    right=4pt,
    top=3pt,
    bottom=3pt,
    breakable,
  },
}

\tikzset{
  >={Latex[length=2mm]},
  every node/.style={font=\sffamily\scriptsize,align=center},
  mod/.style={rounded corners=3pt, draw=Ink, very thick, inner sep=3pt},
  frozen/.style={mod, fill=FrozenBlue},
  train/.style={mod, fill=TrainPeach},
  embed/.style={mod, fill=EmbedGreen},
  dashedbox/.style={draw=Ink, thick, rounded corners=3pt, dashed, inner sep=5pt},
  circ/.style={circle, draw=Ink, fill=white, inner sep=0pt, minimum size=6pt, very thick},
  plus/.style={circle, draw=Ink, fill=white, inner sep=0pt, minimum size=7pt, very thick},
}

\renewcommand{\arraystretch}{1.10}

\theoremstyle{plain}

\theoremstyle{definition}

\theoremstyle{remark}

\usepackage[textsize=tiny]{todonotes}

\title{LumaFlux: Lifting 8-Bit Worlds to HDR Reality with Physically-Guided Diffusion Transformers}

\author[1]{Shreshth Saini}
\author[1]{Hakan Gedik}
\author[2]{Neil Birkbeck}
\author[2]{Yilin Wang}
\author[2]{Balu Adsumilli}
\author[1]{Alan C. Bovik}

\affiliation[1]{The University of Texas at Austin}
\affiliation[2]{Google, Inc.}

\abstract{
The rapid adoption of HDR-capable devices has created a pressing need to convert the 8-bit Standard Dynamic Range (SDR) content into perceptually and physically accurate 10-bit High Dynamic Range (HDR). Existing inverse tone-mapping (ITM) methods often rely on fixed tone-mapping operators that struggle to generalize to real-world degradations, stylistic variations, and camera pipelines, frequently producing clipped highlights, desaturated colors, or unstable tone reproduction. We introduce \textbf{LumaFlux}, a first physically and perceptually guided diffusion transformer (DiT) for SDR-to-HDR reconstruction by adapting a large pretrained DiT. Our LumaFlux introduces (1) a Physically-Guided Adaptation (PGA) module that injects luminance, spatial descriptors, and frequency cues into attention through low-rank residuals; (2) a Perceptual Cross-Modulation (PCM) layer that stabilizes chroma and texture via FiLM conditioning from vision encoder features; and (3) an HDR Residual Coupler that fuses physical and perceptual signals under a timestep- and layer-adaptive modulation schedule. Finally, a lightweight Rational-Quadratic Spline decoder reconstructs smooth, interpretable tone fields for highlight and exposure expansion, enhancing the output of the VAE decoder to generate HDR. To enable robust HDR learning, we curate the first large-scale SDR--HDR training corpus. For fair and reproducible comparison, we further establish a new evaluation benchmark, comprising HDR references and corresponding expert-graded SDR versions. Across benchmarks, LumaFlux outperforms state-of-the-art baselines, achieving superior luminance reconstruction and perceptual color fidelity with minimal additional parameters.
}

\date{\today}
\correspondence{\email{saini.2@utexas.edu}}

\begin{document}


\newcommand*{\vertbar}{\rule[-0.25ex]{0.5pt}{1.5ex}}
\newcommand*{\horzbar}{\rule[.5ex]{2.5ex}{0.5pt}}
\newcommand{\dd}{\mathrm{d}}
\newcommand{\tkernel}{p}
\newcommand{\action}[2]{\left \langle #1, #2\right \rangle }
\newcommand{\bell}{\mathrm{b}}
\newcommand{\norm}[1]
{\left\Vert#1\right\Vert}
\newcommand{\Norm}[1]{\lvert \! \lvert \! \lvert #1 \rvert \! \rvert \! \rvert}
\newcommand{\abs}[1]{\left\vert#1\right\vert}
\newcommand{\babs}[1]{\Big \vert#1 \Big \vert}
\newcommand{\set}[1]{\left\{#1\right\}}
\newcommand{\parr}[1]{\left (#1\right )}
\newcommand{\brac}[1]{\left [#1\right ]}
\newcommand{\ip}[1]{\left \langle #1 \right \rangle }
\newcommand{\Real}{\mathbb R}
\newcommand{\Nat}{\mathbb N}
\newcommand{\Complex}{\mathbb C}
\newcommand{\eps}{\varepsilon}
\newcommand{\too}{\rightarrow}
\newcommand{\bbar}[1]{\overline{#1}}
\newcommand{\wt}[1]{\widetilde{#1}} 
\newcommand{\wh}[1]{\widehat{#1}} 
\newcommand{\diag}{\textrm{diag}} 
\newcommand{\dist}{d} 
\newcommand{\divv}{\mathrm{div}} 
\newcommand{\vol}{\mathrm{vol}} 
\newcommand{\snr}{\mathrm{snr}}
\newcommand{\logsnr}{\rho}
\newcommand{\trace}{\textrm{tr}} 
\def \bfi{\textbf{\footnotesize{i}}} 
\newcommand{\one}{\mathbf{1}}
\newcommand{\zero}{\mathbf{0}}
\newcommand{\vcc}[1]{\mathrm{vec}(#1)}
\newcommand{\mat}[1]{\bm{[} #1 \bm{]}}
\newcommand{\defe}{\coloneqq}

\def \etal{{et al}.}
\newcommand*{\eg}{{\it e.g.}\@\xspace}
\newcommand*{\ie}{{\it i.e.}\@\xspace}

\maketitle



\begin{figure*}[t]
    \centering
    \includegraphics[width=\textwidth]{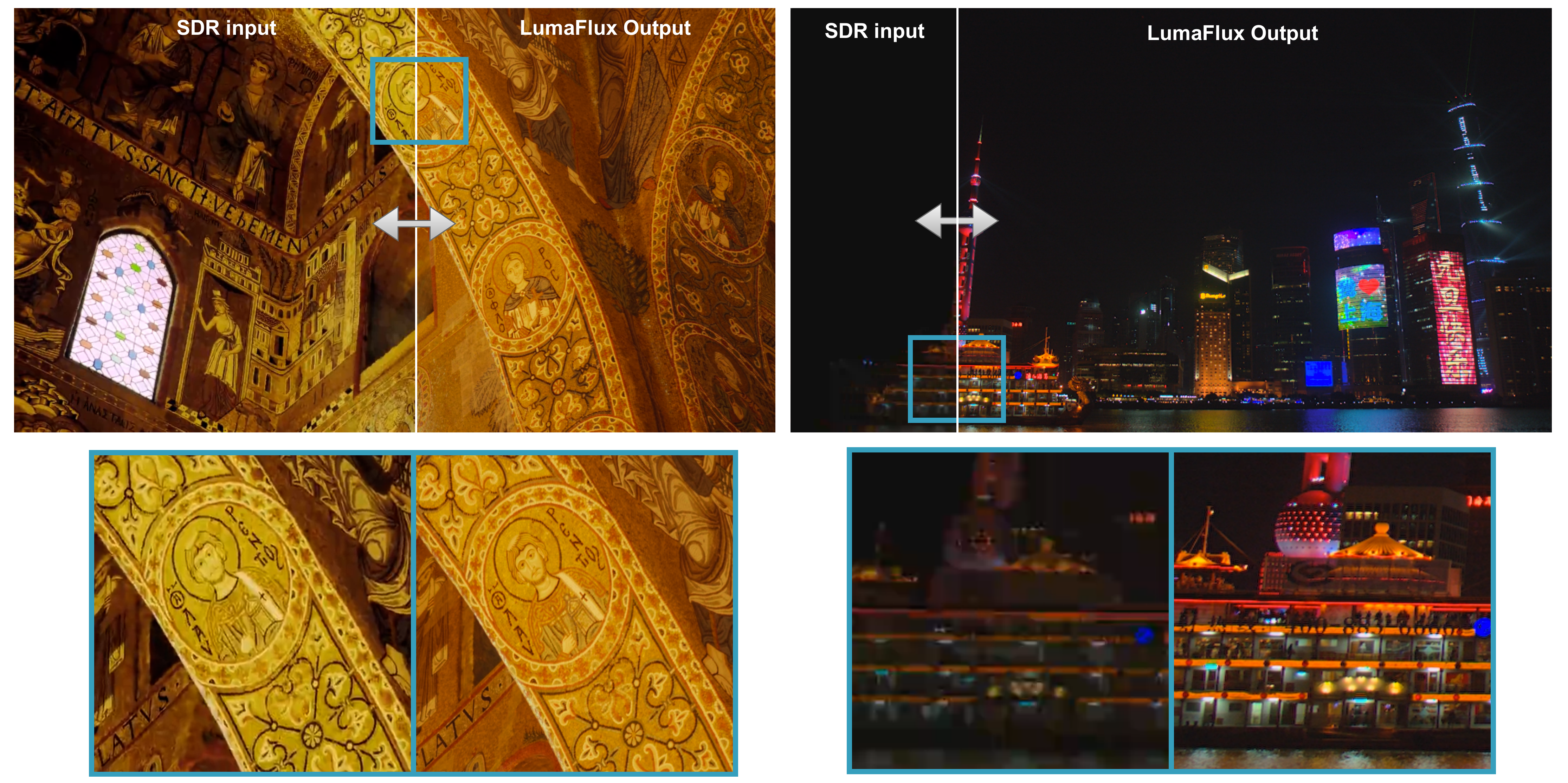}
    \caption{
    We present \textcolor{lumaBlue}{\textbf{LumaFlux}}, a universal Inverse Tone Mapping (ITM) framework that expands dynamic range and color-gamut from 8 bit BT.709 SDR to 10 bit BT.2020 HDR with PQ encoding (i.e. SDRTV-to-HDRTV) of real-world LQ SDR videos. Figure shows examples from Luma-Eval benchmark. Each pair illustrates the SDR$\!\rightarrow$HDR conversion result for one representative frame. LumaFlux restores a substantially broader dynamic range and corrects the color saturation lost during tone-mapping compression, recovering both diffuse texture and specular highlight detail.
    }
    \label{fig:fig-1-compare}
\end{figure*}

\section{Introduction}
\label{sec:intro}

\vspace{-0.3em}
High–Dynamic–Range (HDR) and Wide–Color–Gamut (WCG) imaging have become foundational to modern display and streaming pipelines~\citep{Omnicore2024,CTA2024, ITU2018}. Standards such as ITU-R~BT.2020~\citep{ITU-R_BT.2020} color primaries and the SMPTE~ST~2084~\citep{SMPTE_ST2084} Perceptual Quantizer (PQ) electro–optical transfer function (EOTF) define HDR delivery for consumer displays, enabling up to 10–12-bit precision and luminance peaks exceeding $10^4$~cd/m$^2$. In contrast, most legacy and user-generated content remains confined to 8-bit Standard Dynamic Range (SDR) Rec.709~\citep{ITU-R_BT.709} space, limited to $\approx100$~nits of brightness. Bridging this domain gap: lifting SDR (BT.709) to HDR (PQ, BT.2020) is critical for archival restoration, backward compatibility, and perceptually consistent generative pipelines.

\vspace{0.3em}
\noindent\textbf{From SDRTV to HDRTV.}
The process of converting SDR video into HDR (commonly termed SDRTV→HDRTV or inverse tone mapping (ITM)) seeks to recover the lost dynamic range and colorimetric range of the original scene. Formally, an SDR frame $x_{\mathrm{sdr}}\!\in\![0,1]^3$ is related to its HDR counterpart $x_{\mathrm{hdr}}$ via a forward tone-mapping operator:
\begin{equation}
x_{\mathrm{sdr}} =
\Gamma_{\!\text{OETF}}^{709}
\!\left(
\mathcal{M}_{2020\!\to\!709}
\!\left(
\frac{\Gamma^{2020}_{\text{EOTF}}(x_{\mathrm{hdr}})}{L_{\max}}
\right)
\!\right) + \epsilon,
\label{eq:forward_tm}
\end{equation}
where $\Gamma_{\!\text{OETF}}^{709}$ is the BT.709 opto–electronic transfer function, $\Gamma_{\!\text{EOTF}}^{2020}$ is the BT.2020 electronic–opto transfer function (PQ),
$\mathcal{M}_{2020\!\to\!709}$ represents gamut compression, $L_{\max}$ is the display-referred peak luminance (typically $1000$–$10000$~nits), and $\epsilon$ captures quantization and compression noise. Recovering $x_{\mathrm{hdr}}$ thus requires inverting a nonlinear, device-dependent, and lossy mapping.

\begin{figure}[h]
    \centering
    \includegraphics[width=0.8\textwidth]{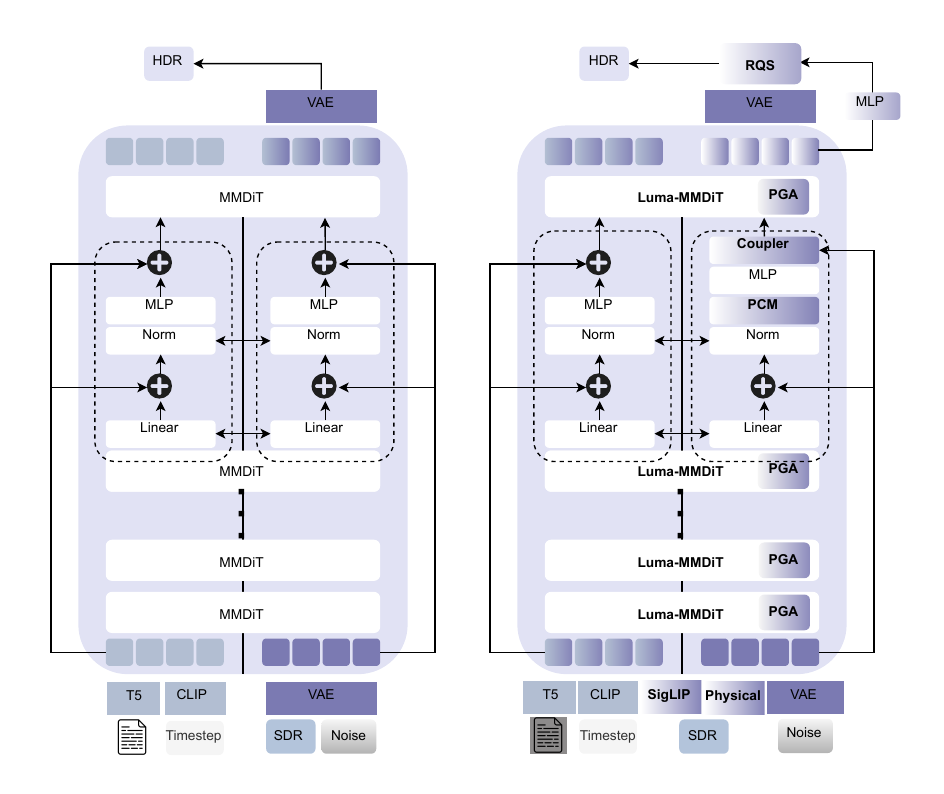}
    \caption{
        \textbf{Architectural Paradigms.}
        \textbf{(Left)} Baseline Diffusion Transformer (DiT) architecture,
        where the backbone can be directly fine-tuned via LoRA or full-weight updates. Such adaptation often overfits small HDR datasets and leads to texture hallucination or unstable luminance restoration. \textbf{(Right)} The proposed \textbf{\textcolor{lumaBlue}{LumaFlux}} introduces lightweight, physically interpretable modules-\textbf{PGA}, \textbf{PCM}, and the \textbf{HDR Residual Coupler}, inserted into the frozen MM-DiT backbone.
        These modules enable prompt-free, physically guided adaptation by modulating attention and MLP activations with luminance, frequency, and perceptual cues. This preserves the pretrained generative prior while allowing accurate ITM with only a few trainable parameters. (Best viewed zoomed in)
    }
    \label{fig:arc_comparison}
\end{figure}

\vspace{0.3em}
\noindent\textbf{Limitations of existing approaches.}
Early ITM methods employed analytical tone–curve models such as Reinhard~\citep{reinhard} or ITU BT.2446~a/b/c~\citep{ITU-R_BT.2407}, which apply global power or knee functions without content awareness. While computationally efficient, these operators cannot reconstruct fine-grained highlight structure, and they produce color shifts when up-converting from BT.709 to BT.2020 due to non-linear interactions between luminance and chroma channels. Learning-based approaches, including CNN and Diffusion Models architectures~\citep{hdrcnn,hdrtv,hdcfm,ictcp,singleHDR, hdrtvdm, jsi-gan, sr-itm, lediff, hdrdm}, learn end-to-end mappings but often overfit to synthetic degradations or fixed tone-mapping datasets~\citep{hdrtv, hdrtvdm}, limiting generalization to real SDR content with compression artifacts or exposure bias. Recent diffusion-based methods such as LEDiff~\citep{lediff} leverage generative priors but focus on alternate exposure fusion~\citep{hu2024generating, zhou2025flow}, require full model retraining, large-scale paired data, and text prompts that can introduce semantic drift during generation.

\vspace{0.3em}
We propose \textcolor{lumaBlue}{\textbf{LumaFlux}}, a physically and perceptually guided diffusion transformer that performs SDR$\!\rightarrow$HDR reconstruction by adapting a large pretrained Flux~\citep{flux2024} backbone without retraining its weights.
Instead of direct regression or prompt-based fine-tuning (see Fig.~\ref{fig:arc_comparison}), LumaFlux inserts time- and layer-adaptive adapters that inject physically grounded luminance cues and perceptual semantics in a parameter-efficient manner. It comprises three synergistic modules:
(1)~a Physically-Guided Adaptation (PGA) layer encoding luminance, spatial descriptors, and spectral features through gated low-rank attention updates;
(2)~a Perceptual Cross-Modulation (PCM) block applying FiLM~\citep{film} conditioning from frozen SigLIP~\citep{siglip} embeddings to preserve texture and color constancy; and
(3)~an HDR Residual Coupler that fuses both pathways using timestep- and layer-dependent scaling.
Finally, a lightweight Rational-Quadratic Spline (RQS) decoder ensures smooth, invertible tone expansion for physically plausible highlight reconstruction (see Fig.~\ref{fig:fig-1-compare}).

To enable scalable HDR training, we curate the first large-scale SDR–HDR training corpus by combining HDR source videos from HIDROVQA~\citep{hidrovqa}, CHUG~\citep{chug}, and LIVE-TMHDR~\citep{livetmhdr}. All sources are normalized to PQ-encoded BT.2020 color space, ensuring consistent physical luminance representation. We further introduce \textbf{Luma-Eval}, a new evaluation benchmark for ITM consisting of HDR references and corresponding expert graded SDR versions, facilitating reproducible cross-method comparison.
 Our main contributions are summarized as follows:
\begin{itemize}
    \item We propose \textcolor{lumaBlue}{\textbf{LumaFlux}}, the first physically and perceptually guided diffusion transformer that performs ITM by adapting a large pretrained diffusion model.
    \item We design three lightweight modules PGA, PCM, and HDR Residual Coupler that align physical luminance modeling with perceptual color semantics under timestep and layer adaptive control.
    \item We curate the first large-scale SDR–HDR training corpus and introduce a new evaluation benchmark, Luma-Eval, unified in PQ–Rec.2020 space.
    \item Extensive experiments confirm that LumaFlux achieves state-of-the-art fidelity and perceptual realism across multiple benchmarks, all in a prompt-free and computationally efficient design.
\end{itemize}

\section{Related Work}
\label{sec:related}

\vspace{-0.3em}
\noindent\textbf{Classical inverse tone mapping.}
Early efforts in ITM relied on physically inspired tone expansion curves. Pioneering operators such as Reinhard~\citep{reinhard}, and the family of ITU BT.2446~a/b/c~\citep{ITU-R_BT.2407} apply analytic luminance mappings (\eg, knee, shoulder, or logarithmic functions) coupled with global saturation control. While computationally efficient, these methods lack content adaptivity and tend to over-brighten low-exposure regions or clip high-intensity highlights. Moreover, they are inherently limited by the non-invertibility of the forward tone-mapping chain that includes opto-electronic transfer functions (OETFs), gamut compression, and quantization. Consequently, such deterministic mappings struggle when converting real 8-bit SDR Rec.709 video to 10-bit HDR Rec.2020, which demands both dynamic range and wide-gamut consistency.

\vspace{0.3em}
\noindent\textbf{Learning-based SDR$\!\rightarrow$HDR reconstruction.}
With the rise of deep learning, data-driven ITM models have replaced hand-crafted operators. CNN-based methods such as HDRCNN~\citep{hdrcnn}, Deep SR-ITM~\citep{sr-itm}, and HDRTVNet~\citep{hdrtv} formulate SDR$\!\rightarrow$HDR as a supervised regression task over paired SDR–HDR frames. To improve spatial adaptivity, HDCFM~\citep{hdcfm} and HDRTVDM~\citep{hdrtvdm} introduce hierarchical feature modulation and dynamic context transformation. However, these convolutional networks are limited by local receptive fields and struggle to generalize to unseen degradations, especially compression noise and mixed tone curves commonly found in user-generated content (UGC).
They also operate in fixed color spaces, often Rec.709, without physically consistent transfer to PQ or Rec.2020 domains. Transformer-based methods have recently extended receptive fields and global modeling capacity. HDRTransformer~\citep{hdrtransformer} exploits self-attention for long-range correlation, but remain fully supervised and use exposure fusion instead of direct HDR prediction. In addition, most prior datasets (\eg, HDRTV1K~\citep{hdrtv}, HDRTV4K~\citep{hdrtvdm}) are either synthetic or tone-mapped from HDR masters using fixed operators, leading to narrow domain diversity.

\vspace{0.3em}
\noindent\textbf{Diffusion and generative-model approaches.}
Diffusion models~\citep{ho2020ddpm,rombach2022high} have demonstrated strong priors for image generation and restoration. Recent work such as LEDiff~\citep{lediff} leverages latent diffusion to expand LDR dynamic range by fusing exposure cues within the latent space, while HDRDM~\citep{hdrdm} utilizes diffusion priors to mitigate highlight artifacts. Although effective, these models require retraining of large backbones or rely on text prompts for conditioning, or does the exposure fusion instead of true HDR prediciton, which introduces semantic drift, limits the dynamic range expansion, and limits their applicability to uncaptioned video frames. Moreover, they lack explicit physical alignment with luminance transfer functions, making highlight reconstruction perceptually inconsistent.


\vspace{0.3em}
\noindent\textbf{HDR datasets and evaluation benchmarks.}
The progress of HDR reconstruction has been driven by curated datasets such as HDRTV1K~\citep{hdrtv} and HDRTV4K~\citep{hdrtvdm}, each providing paired SDR–HDR frames derived via controlled tone mapping. However, these datasets lack the diversity and real-capture noise found in consumer SDR videos. The LIVE-TMHDR dataset~\citep{livetmhdr}, originally developed for HDR tone-mapping quality assessment, contains 40 HDR source videos and 10 tone-mapped SDR versions including expert tone mapping, acts as the ideal candidate for our task. In this work, we curate LIVE-TMHDR and integrate it with HIDROVQA~\citep{hidrovqa} and CHUG~\citep{chug} to form the first large-scale SDR–HDR corpus, unified under PQ-encoded BT.2020 representation. We also propose a new evaluation benchmark, LumaEval, establishing a high-fidelity reference for fair, perceptual HDR reconstruction assessment.

\vspace{0.3em}
\noindent\textbf{Summary.}
In contrast to previous approaches, \textbf{\textcolor{lumaBlue}{LumaFlux}} is the first framework to adapt a large pretrained diffusion transformer for physically grounded SDR$\!\rightarrow$HDR conversion without textual prompts or backbone retraining.
It bridges luminance modeling, physical statistics, and perceptual diffusion priors,
achieving state-of-the-art performance across both synthetic and real benchmarks.

\section{Preliminaries}
\label{sec:prelim}

\vspace{-0.3em}
\noindent
This section outlines the fundamentals of ITM and the generative transformer backbone underlying \textcolor{lumaBlue}{\textbf{LumaFlux}}. We first describe the forward tone-mapping process converting HDR signals to SDR,  then summarize the diffusion–transformer formulation used in modern latent generative models (\eg, Flux~\citep{flux2024}, SD3~\citep{sd32024}).

\vspace{0.3em}
\noindent
\textbf{Problem definition.}
Given an HDR frame $x_{\mathrm{hdr}}$ expressed in absolute luminance and wide-gamut chromaticity, a standard camera or broadcast pipeline compresses it to an 8-bit SDR frame $x_{\mathrm{sdr}}$ using a tone-mapping and encoding process~\eqref{eq:forward_tm}. The goal of inverse tone mapping (ITM) is to approximate the inverse of~\eqref{eq:forward_tm}:
\begin{equation}
\widehat{x}_{\mathrm{hdr}} = f_{\boldsymbol{\theta}}(x_{\mathrm{sdr}}),
\label{eq:itm_goal}
\end{equation}
where $f_{\boldsymbol{\theta}}$ reconstructs a perceptually faithful 10-bit HDR representation in the PQ (SMPTE~ST~2084) BT.2020 domain. This task is ill-posed: tone-mapping is nonlinear, many-to-one, and discards scene-referred luminance details. Recovering these lost degrees of freedom requires both \emph{physical priors} (luminance and color constraints) and \emph{perceptual priors} (semantics and realism) learned from large-scale generative models, motivating our use of diffusion–transformer backbones.

\subsection{Diffusion Transformers}
\label{subsec:dits}

\vspace{-0.3em}
\noindent
Recent diffusion transformers such as SD3~\citep{sd32024} and Flux~\citep{flux2024} use the Multimodal Diffusion Transformer (MM-DiT) as their core unit, unifying flow-matching dynamics with multimodal attention for cross-domain conditioning across visual, textual, and auxiliary signals.

\vspace{0.3em}
\noindent
\textbf{Continuous generative formulation.}
The generative process can be expressed as a continuous-time ordinary differential equation (ODE) or stochastic differential equation (SDE) defined over latent tokens $\mathbf{z}_t$:
\begin{equation}
\frac{d\mathbf{z}_t}{dt} = -\mathcal{F}_{\boldsymbol{\theta}}(\mathbf{z}_t, t, \mathbf{c}),
\label{eq:diffusion}
\end{equation}
where $\mathbf{c}$ denotes conditioning signals (e.g., text or auxiliary image embeddings) and $\mathcal{F}_{\boldsymbol{\theta}}$ is a time-conditioned transformer predicting the instantaneous velocity field. Flow-matching~\citep{lipman2023flow,liu2023flow, saini2025rectified} ensures $\mathcal{F}_{\boldsymbol{\theta}}$ approximates the optimal transport between prior and data distributions.

\vspace{0.3em}
\noindent
\textbf{Transformer block structure.}
Each MM-DiT layer refines the latent representation $\mathbf{z}^{\ell}\!\in\!\mathbb{R}^{N\times d}$ via multi-head self-attention and feed-forward modulation:
\begin{equation}
\mathbf{z}^{\ell+1} =
\mathbf{z}^{\ell}
+
\mathrm{MSA}\!\big(\mathrm{LN}(\mathbf{z}^{\ell})\big)
+
\mathrm{MLP}\!\big(\mathrm{LN}(\mathbf{z}^{\ell})\big),
\end{equation}
where $\mathrm{MSA}$ denotes multi-head self-attention and $\mathrm{MLP}$ the feed-forward network. Each attention head computes:
\begin{equation}
\mathrm{Attn}(\mathbf{Q}, \mathbf{K}, \mathbf{V}) =
\mathrm{softmax}\!\left(\frac{\mathbf{Q}\mathbf{K}^\top}{\sqrt{d_k}}\right)\mathbf{V}.
\end{equation}
In MM-DiTs, $\mathbf{Q}$ arises from visual tokens $\mathbf{z}_t$, while $\mathbf{K}$ and $\mathbf{V}$ may originate from text, image, or other perceptual embeddings, forming a unified multimodal context.

\begin{tcolorbox}[lumaBoxGreen,title={Notation Highlights}]
$\mathcal{M}_{709\!\leftrightarrow\!2020}$ – Color-space transform matrices.\\
$\Gamma^{709}_{\!\text{OETF}}$, $(\Gamma^{\text{PQ}}_{\!\text{EOTF}})^{-1}$ – SDR/HDR transfer functions.\\
$x_{\mathrm{sdr}}$, $x_{\mathrm{hdr}}$ – SDR and HDR frames.\\
$\mathbf{z}_t$, $\mathcal{F}_{\boldsymbol{\theta}}$ – Latent state and diffusion transformer.\\
$\lambda_t^{\ell}$ – Timestep/layer modulation coefficient.\\
\textcolor{lumaBlue}{$\phi_{\mathrm{pga}}$, $\phi_{\mathrm{pcm}}$} – Physical and perceptual adapters used in LumaFlux.
\end{tcolorbox}


\begin{figure*}[t]
    \centering
    \includegraphics[width=\textwidth]{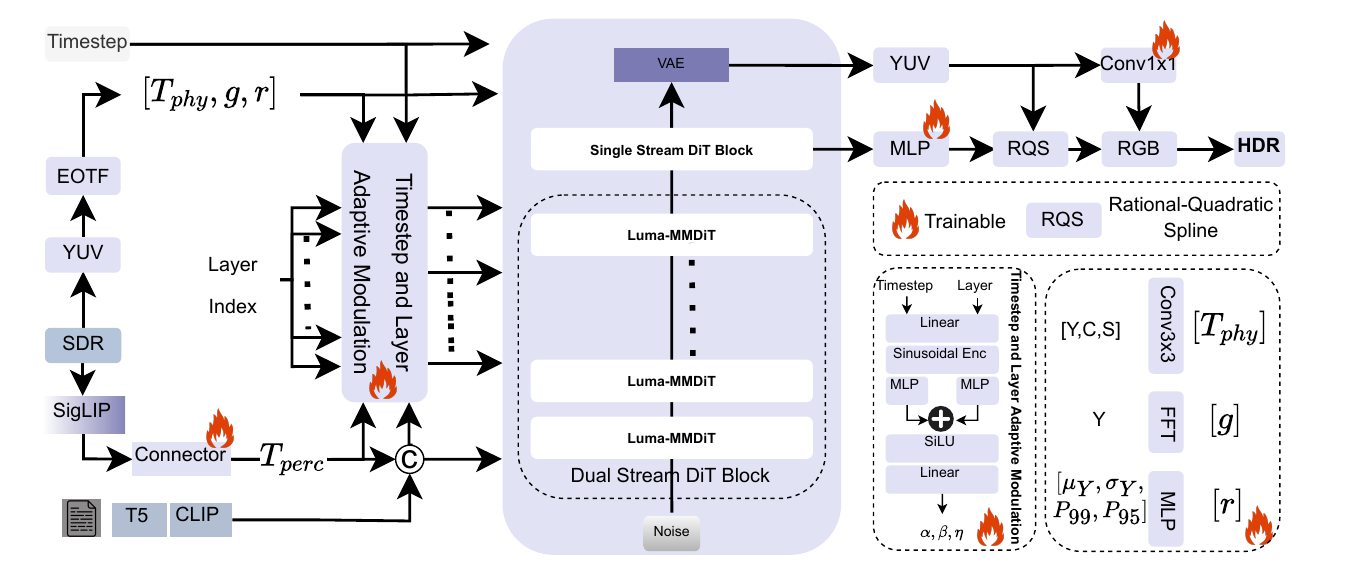}
    \caption{\textbf{LumaFlux Overview.}
      SDR input is split into physical (\(T_{\text{phys}}\)) and perceptual (\(T_{\text{perc}}\)) streams. With timestep/layer conditioning \(\Psi(t,\ell)\), In each Luma-MMDiT block (see Fig.~\ref{fig:arc_comparison}), PGA injects luminance and spectrum aware LoRA updates into attention, PCM applies FiLM~\citep{film} to modulate normalized features, and an HDR Residual Coupler fuses both cues. A frozen VAE decoder with an RQS tone-field head reconstruct HDR in PQ/BT.2020.
    }
    \label{fig:method_overview}
\end{figure*}

\section{Method}
\label{sec:method}

\vspace{-0.3em}
\noindent
We present \textcolor{lumaBlue}{\textbf{LumaFlux}}, a physically and perceptually guided adaptation framework on a frozen Flux~\citep{flux2024} MM-DiT backbone. LumaFlux reconstructs HDR from 8-bit SDR by inserting parameter-efficient adapters that (i) inject physical luminance structure and (ii) preserve perceptual semantics, all while preserving the pretrained generative manifold. As shown in Fig.~\ref{fig:method_overview}, the trainable components are:
(1)~Physically-Guided Adaptation (PGA),
(2)~Perceptual Cross-Modulation (PCM),
(3)~an HDR Residual Coupler, and
(4)~a Rational–Quadratic–Spline (RQS) tone-field decoder.
All backbone weights remain frozen.

\subsection{Feature Extraction and Conditioning}

\vspace{-0.3em}
\noindent
\textbf{Physical features.}
We first obtain input frame linear light in BT.2020 using PQ EOTF.
Let \(Y=\mathbf{m}_{2020}^{\!\top}x_{\mathrm{lin}}\) with \(\mathbf{m}_{2020}=[0.2627,\,0.6780,\,0.0593]^\top\). We assemble local physical descriptors:

\begin{align}
T_{\mathrm{phys}}
&= \mathrm{Conv}_{3\times 3}\!\left(
[\, Y,\ \log(1 + |\nabla Y|),\ \mathrm{sat} \,]
\right),
\end{align}

and global stats \(s_g=[\mu_Y,\sigma_Y,p_{95},p_{99}]\) summarized by \(g=\mathrm{MLP}_g(s_g)\).

\textbf{Perceptual features.}
We extract semantic embeddings with a frozen SigLIP image encoder~\citep{siglip}:
\begin{align}
T_{\mathrm{perc}} \;=\; \phi_{\mathrm{SigLIP}}(x_{\mathrm{sdr}})\ \in \mathbb{R}^{N_{\!p}\times d_p}.
\end{align}
A learned connector \(C_{\mathrm{perc}}\) projects to the MM-DiT channel size to enable token-level modulation inside the backbone, and also concatenated with prompt tokens for semantic details.

\textbf{Spectral descriptor.}
We compute a luminance spectrum from \(Y\) using a single FFT~\footnote{We use Pytorch's rfft2 to compute compact fourier transform} pass and energy pooling into \(K\) bands:
\(
r \in \mathbb{R}^{K}
\).
This descriptor modulates high-frequency restoration via spectral gating (Sec.~\ref{subsec:pga}).

\subsection{Timestep$–$Layer Adaptive Modulation}
\label{subsec:modulation}

\vspace{-0.3em}
\noindent
A shared conditioning block \(\Psi(t,\ell)\), visualized in Fig.~\ref{fig:method_overview}, produces all block-wise modulation parameters:

\begin{align}
\bigl[\,
\alpha_{\mathrm{pga}}^{t,\ell},\
\beta_{\mathrm{pga}}^{t,\ell},\
\alpha_{\mathrm{pcm}}^{t,\ell},\
\beta_{\mathrm{pcm}}^{t,\ell},\
n_{\mathrm{spec}}^{t,\ell},\
\lambda^{t,\ell}
\,\bigr]
&= \Psi(t,\ell).
\end{align}

where \(t\) is the flow time and \(\ell\) the block index. We encode \((t,\ell)\) with learned linear layers plus sinusoidal features, fuse them by addition, apply \(\mathrm{SiLU}\), then output the parameters through affine heads. Early layers/large \(t\) receive stronger global tone gains; late layers/small \(t\) focus on highlight details.

\subsection{Physically-Guided Adaptation (PGA)}
\label{subsec:pga}

\vspace{-0.3em}
\noindent
PGA injects physically grounded luminance cues into attention via a gated low-rank LoRA~\citep{lora}. Unlike standard LoRA, it conditions attention heads on luminance, gradients, saturation, and spectral energy, adaptively enhancing highlights and textures only where needed. Dynamic gates $\mathbf{G}{\mathrm{phys}}$ and spectral terms $g{\mathrm{FFT}}$ regulate frequency and intensity, preserving highlight contrast while preventing over-expansion in smooth areas yielding energy-consistent tone reconstruction and stable HDR recovery.

Let \(\mathbf{W}^{(0)}_V\) be a frozen value-projection. We add a gated low-rank residual:
\begin{equation}
R_v^{\text{base}} \;=\; \mathbf{A}_v\mathbf{B}_v \ \in\ \mathbb{R}^{d\times d},
\qquad \mathrm{rank}(R_v^{\text{base}})=r\ll d,
\end{equation}
and compute per-head gates from physical cues:
\begin{align}
\mathbf{G}_{\mathrm{phys}}
&= \mathrm{Diag}\!\left(\sigma\!\big(P_v\,[\,T_{\mathrm{phys}}\,\Vert\,g\,]\big)\right), \\
g_{\mathrm{FFT}}
&= \mathrm{softplus}(W_r\,r)\ \in\mathbb{R}^{H_{\text{attn}}},
\end{align}
where \(P_v,W_r\) are $1{\times}1$ pointwise projections to head dimensions, \(\sigma\) is sigmoid, and \(H_{\text{attn}}\) is the number of heads. Timestep/layer-adaptive scaling and spectral gating yield:

\begin{equation}
\begin{split}
R_v^{t,\ell} =\;
&(\alpha_{\mathrm{pga}}^{t,\ell} R_v^{\text{base}}
+ \beta_{\mathrm{pga}}^{t,\ell}\mathbf{I}) \,
\mathbf{G}_{\mathrm{phys}} \\
&\times \big(\mathbf{I}
+ n_{\mathrm{spec}}^{t,\ell}\mathrm{Diag}(g_{\mathrm{FFT}})\big)
\end{split}
\label{eq:pga_final}
\end{equation}

and the attention value projection is updated as
\(
\mathbf{W}_V \leftarrow \mathbf{W}^{(0)}_V + R_v^{t,\ell}.
\)
\(T_{\text{phys}}\) and \(r\) strengthen high-intensity and high-frequency pathways only when the scene requires them, preventing overexpansion in flat regions and improving highlight roll-off fidelity.

\subsection{Perceptual Cross-Modulation (PCM)}
\label{subsec:pcm}

\vspace{-0.3em}
\noindent
While the PGA stream focuses on physically grounded luminance cues, many tone-mapping distortions are perceptual, arising from the loss of semantic and chromatic coherence across spatial regions. Direct low-level restoration (e.g., L1 or gradient-based) fails to recover fine perceptual relationships. Perceptual Cross-Modulation (PCM) solves this by conditioning hidden transformer states on SigLIP embeddings~\citep{siglip}, using feature-wise modulation~\citep{film,chen2022adaptformer,zhao2024dynamic,choidiffusion} to rescale and shift normalized activations. This enforces color constancy, semantic coherence, and stable tone across illumination and content variations.

Let $\mathbf{h}^{\ell}$ be the post-normalization hidden activation in an MM-DiT block.
A small MLP first projects SigLIP embeddings $T_{\mathrm{perc}}$ into the same latent width via a connector $C_{\mathrm{perc}}$:
\[
T_{\mathrm{perc}}' = C_{\mathrm{perc}}(T_{\mathrm{perc}}) \in \mathbb{R}^{N\times d}.
\]
The modulation coefficients are then computed as
\begin{align}
[\gamma^{t,\ell},\ \zeta^{t,\ell}]
&= \alpha_{\mathrm{pcm}}^{t,\ell}\,\mathrm{MLP}\!\big(T_{\mathrm{perc}}'\big) + \beta_{\mathrm{pcm}}^{t,\ell},\\
\mathrm{PCM}(\mathbf{h}^{\ell})
&= \gamma^{t,\ell}\odot \mathrm{LN}(\mathbf{h}^{\ell}) + \zeta^{t,\ell},
\end{align}
where $\odot$ denotes element-wise scaling and $\mathrm{LN}$ is layer normalization.
Applying FiLM~\citep{film} after normalization ensures scale-invariant conditioning that remains stable across brightness levels and diffusion timesteps. This improves both spatial and inter-frame perceptual continuity, analogous to AdaptFormer~\citep{chen2022adaptformer} and T2I-Adapter~\citep{mou2024t2i}, but in a parameter-efficient FiLM form.

\subsection{HDR Residual Coupler}
\label{subsec:coupler}

\vspace{-0.3em}
\noindent
We fuse physical and perceptual residuals, that flow through the architecture similar to $z^0$ residual latent, with a time/layer-gated path:
\begin{equation}
{\;
\mathbf{z}_{\mathrm{out}}^{\ell} \;=\;
\mathbf{z}_{\mathrm{res}}^{\ell} \;+\;
\lambda_{t}^{\ell}\Big(W_p\,T_{\mathrm{phys}} + W_c\,C_{\mathrm{perc}}(T_{\mathrm{perc}})\Big)
\;}
\label{eq:coupler_full}
\end{equation}
where \(\mathbf{z}_{\mathrm{res}}^{\ell}\) is the MMDiT block residual output, and \(W_p,W_c\) are \(1{\times}1\) projections into the token dimension. \(\lambda_{t}^{\ell}\) decays low-frequency luminance corrections as \(t\!\downarrow\!0\), letting late layers focus on fine highlight structure. This residual formulation functions as a form of dynamic guidance flow, similar in spirit to classifier-free guidance but implemented through additive couplings within the latent manifold. Early timesteps with large $\lambda_t^{\ell}$ prioritize global tone recovery (contrast and exposure alignment), whereas smaller $\lambda_t^{\ell}$ at later steps refine texture and local highlight roll-offs. The coupling therefore acts as a temporal–spatial gate that balances physical luminance accuracy with perceptual realism throughout the diffusion trajectory.

\subsection{Rational$–$Quadratic Spline (RQS) Tone$-$Field Decoder}
\label{subsec:rqs}

\vspace{-0.3em}
\noindent
Recovering accurate HDR luminance from a frozen decoder is non-trivial since the underlying VAE in Flux~\citep{flux2024} is trained for 8-bit SDR distributions. Direct regression often leads to highlight clipping or saturation collapse. To address this, we introduce a lightweight, learnable Rational–Quadratic Spline (RQS)~\citep{rqs} tone-field that locally expands and reshapes luminance while ensuring monotonicity and differentiability, crucial for physically plausible tone reconstruction and stable gradient propagation during training.

A frozen VAE decoder \(\mathcal{D}_{\mathrm{VAE}}\) yields a preliminary frame \(x_{\mathrm{out}}=\mathcal{D}_{\mathrm{VAE}}(\mathbf{z}_T)\). We convert to YUV (BT.2020) to obtain \(Y_{\mathrm{out}}\). A lightweight MLP head predicts monotone spline parameters \((\boldsymbol{\xi},\boldsymbol{\eta},\boldsymbol{s})\)~\citep{rqs}:
\begin{equation}
\widehat{Y} \;=\; \mathrm{RQS}\!\big(Y_{\mathrm{out}};\,\boldsymbol{\xi},\boldsymbol{\eta},\boldsymbol{s}\big),
\end{equation}
while \(\widehat{U},\widehat{V}\) are refined via \(\mathrm{Conv}_{1\times1}\). Unlike polynomial or exponential tone curves, RQS~\citep{rqs} provides locally adaptive, differentiable, and invertible tone expansion with bounded derivatives, producing smooth highlight knees without banding, allowing LumaFlux to reconstruct highlight roll-offs and low-light gradients that are typically lost in SDR-trained decoders. Final output is given as:

\begin{align}
\widehat{x}_{\mathrm{hdr}} = \mathcal{M}_{\mathrm{YUV}\!\to\!\mathrm{RGB}}([\widehat{Y},\widehat{U},\widehat{V}])_{\mathrm{PQ,\,BT.2020}}.
\end{align}

\vspace{1em}
\noindent
\textbf{Training Objective}
Our loss function consists of three terms, which jointly supervise both physical and perceptual fidelity. Given HDR reference \(x^\star\) (PQ/BT.2020) and our output \(\widehat{x}\):
{\small
\begin{equation}
\mathcal{L} =
\lambda_1\!\big\|Y_{\mathrm{lin}}{-}Y_{\mathrm{lin}}^\star\big\|_1
+ \lambda_2\!\big\|x_{\mathrm{lin}}{-}x_{\mathrm{lin}}^\star\big\|_1
+ \lambda_3\,\mathcal{L}_{\mathrm{spline\mbox{-}smooth}}
\label{eq:training_loss_full}
\end{equation}
}

\noindent
where linear terms use inverse PQ EOTF; \(\mathcal{L}_{\mathrm{spline\mbox{-}smooth}}\) penalizes adjacent RQS knot slope changes (stabilizes tone continuity).



\section{Large-Scale Real-World High-Quality SDR–HDR Dataset for ITM}
\label{sec:dataset}

\vspace{-0.3em}
\noindent
Training and evaluating inverse tone mapping (ITM) models demands datasets that
accurately capture the physical properties of real HDR imagery as well as the nonlinear degradations that occur in practical SDR production pipelines. Existing benchmarks are limited: HDRTV1K~\citep{hdrtv} relies on YouTube's SDR approximations from fixed tone mappers, while HDRTV4K~\citep{hdrtvdm} provides limited dynamic diversity. Consequently, models trained on such data tend to overfit specific tone styles and fail to generalize to real-world SDR content with mixed camera pipelines and compression. To enable physically grounded training and fair evaluation, we curate the first unified PQ-normalized dataset for real-world SDR$\!\rightarrow$HDR restoration, spanning professional (PGC) and user-generated (UGC) HDR sources, multiple tone-mapping operators including expert graded SDR, and codec degradations.

\vspace{0.3em}
\noindent
\textbf{Training Dataset Construction.}
Our training corpus combines three complementary HDR sources with diverse capture conditions. HIDROVQA~\citep{hidrovqa} provides 411 HDR reference PGC videos used in VQA research, covering diverse content and tone scales. CHUG~\citep{chug} provides 428 crowdsourced HDR UGC videos with handheld motion, exposure bias, and typical mobile HDR noise characteristics. LIVE-TMHDR~\citep{livetmhdr} provides 40 studio-quality (PGC) HDR reference videos with SDR counterparts, including\emph{expert tone-mapped} SDR for perceptual ground truth. To prevent domain imbalance, we sample PGC and UGC HDR content in a 1:1 ratio during training. Each HDR frame is paired with multiple SDR variants generated by a \emph{composite degradation chain}:
\begin{equation}
x_{\mathrm{sdr}} =
\mathcal{Q}_{\text{codec}}
\!\circ\!
\mathcal{M}_{2020\!\to\!709}
\!\circ\!
\mathcal{TMO}
\!\left(x_{\mathrm{pq}}; \theta_{\mathrm{tone}}\right),
\label{eq:degradation_chain}
\end{equation}
where $\mathcal{TMO}$ denotes a tone-mapping operator (e.g., Reinhard, BT.2446c, YouTube-LogC), $\mathcal{M}_{2020\!\to\!709}$ applies gamut compression to SDR primaries, and $\mathcal{Q}_{\text{codec}}$ models quantization and compression. We use same TMOs as suggested in ~\citep{hdrtvdm}, additionally, we employ CRF values of 23, 31, and 39 to simulate high, medium, and low-quality SDR encodings.

\noindent
\textbf{Sampling strategy.}
To achieve scale while maintaining diversity, we sample at varying frame rates per source:
\begin{itemize}[leftmargin=1.3em]
\item Expert-tone-mapped SDR (LIVE-TMHDR~\citep{livetmhdr}): 60~fps sampling, 10s average duration, yielding $\!\approx\!600$ frames/video for 30 training and 10 evaluation videos, $\!\approx\!54\mathrm{k}$ pairs (with 3~CRF levels).
\item Synthetic TMO variants (all sources): 10~fps remaining LIVE-TMHDR~\citep{livetmhdr}, and 1~fps all PGC~\citep{hidrovqa} and UGC~\citep{chug} HDR, yielding $\!\approx\!9\mathrm{k}$ to $\!\approx\!12\mathrm{k}$ frame pairs per TMO.
\end{itemize}
Combining 8~TMOs with compression levels yields $\!\approx\!264\mathrm{k}$ SDR–HDR pairs,
plus an additional 54k expert-tone-mapped reference pairs. This provides both scale and stylistic diversity needed for robust inverse tone mapping.

\vspace{-0.3em}
\noindent
\textbf{Evaluation Benchmark.}
For standardized comparison, we create a unified evaluation benchmark that merges
existing datasets and introduces a new subset:
\textbf{Luma-Eval}.
It includes both PGC and UGC sources under consistent PQ BT.2020 encoding as done for training dataset. Specifically, the benchmark comprises:
(1)~HDRTV1K~\citep{hdrtv},
(2)~HDRTV4K~\citep{hdrtvdm}, and
(3)~\textbf{Luma-Eval}—a newly curated set of 20 HDR source videos (10 from LIVE-TMHDR~\citep{livetmhdr} and 10 from CHUG~\citep{chug}) unseen during training.
Benchmark includes:
ground-truth HDR reference,
expert tone-mapped SDR (for PGC content), and
8~synthetic SDR variants across 3~CRF compression levels.
This ensures evaluation under both perceptually optimized and degradation-heavy SDR conditions.

\noindent
\textbf{HDR Normalization and Color Standardization.}
All HDR sources are standardized to the \textit{PQ-encoded BT.2020} color domain,
ensuring consistent luminance and chromatic representation across heterogeneous datasets.
For each HDR frame $x_{\mathrm{hdr}}^{(i)}$ with primaries $p_i$ (e.g., P3, ACEScg)
and native transfer function $\Gamma_i$, we compute its normalized counterpart:
\begin{equation}
x_{\mathrm{pq}}^{(i)} =
\Gamma_{\!\text{OETF}}^{\mathrm{PQ}}
\!\left(
\mathcal{M}_{p_i\!\to\!2020}
\!\big((\Gamma_i)^{-1}(x_{\mathrm{hdr}}^{(i)})\big)
\right),
\label{eq:pq_norm}
\end{equation}
where $\mathcal{M}_{p_i\!\to\!2020}$ performs gamut conversion to BT.2020 primaries.
Linear luminance values are clipped to $10^4$~cd/m$^2$, consistent with HDR mastering metadata. This step eliminates ambiguity in color encoding and guarantees that supervision corresponds to physically meaningful HDR recovery.

\begin{tcolorbox}[colback=lumaOrange!5,colframe=lumaOrange!80!black,title={Benchmark Standardization}]
All evaluation samples are stored in PQ BT.2020 at 10-bit depth with identical mastering metadata (1,000~nits peak luminance), ensuring consistent physical calibration across datasets.
\end{tcolorbox}

\begin{table*}[t]
\centering
\tiny
\setlength{\tabcolsep}{1.7pt}
\renewcommand{\arraystretch}{1.3}
\caption{
\textbf{Quantitative comparison across benchmarks.}
We compare \textbf{\textcolor{lumaBlue}{LumaFlux}} with CNN-based and generative methods. Metrics are computed in PU21 space for luminance (\textbf{PSNR}, \textbf{SSIM}), HDR domain (\textbf{HDR-VDP3}), and perceptual color ($\mathbf{\Delta E_{\mathrm{ITP}}}$, lower is better).
\textbf{\textcolor{lumaBlue}{LumaFlux}} achieves the best overall performance across datasets, consistently improving both dynamic range and colorimetric fidelity.
Best results are in \textbf{bold}, second best are \underline{underlined}.
}
\label{tab:all_benchmarks}
\begin{tabular}{lcccccccccccc}
\toprule
\multirow{2}{*}{\textbf{Method}}
& \multicolumn{4}{c}{\textbf{HDRTV1K}}
& \multicolumn{4}{c}{\textbf{HDRTV4K}}
& \multicolumn{4}{c}{\textbf{Luma-Eval}} \\
\cmidrule(lr){2-5} \cmidrule(lr){6-9} \cmidrule(lr){10-13}
& PSNR$\uparrow$ & SSIM$\uparrow$ & HDR-VDP3$\uparrow$ & $\Delta E_{\mathrm{ITP}}\downarrow$
& PSNR$\uparrow$ & SSIM$\uparrow$ & HDR-VDP3$\uparrow$ & $\Delta E_{\mathrm{ITP}}\downarrow$
& PSNR$\uparrow$ & SSIM$\uparrow$ & HDR-VDP3$\uparrow$ & $\Delta E_{\mathrm{ITP}}\downarrow$ \\
\midrule
HDRTVNet++~\citep{hdrtv} & 38.36 & 0.973 & 8.75 & 8.28 & 30.82 & 0.881 & 8.12 & 7.85 & 36.54 & 0.901 & 8.22 & 7.35 \\
ICTCPNet~\citep{ictcp} & 36.59 & 0.922 & 8.57 & 7.79 & 33.12 & 0.977 & 8.90 & \underline{6.75} & 34.45 & 0.919 & 8.74 & 6.93 \\
HDRTVDM~\citep{hdrtvdm} & 36.98 & 0.971 & 8.55 & 10.84 & 30.15 & 0.886 & 7.90 & 9.95 & 35.10 & 0.903 & 8.24 & 9.44 \\
HDCFM~\citep{hdcfm} & \underline{38.42} & \underline{0.973} & \underline{8.52} & \underline{7.83} & 33.25 & 0.908 & 8.20 & 7.42 & \underline{36.78} & \underline{0.915} & 8.29 & 7.20 \\
Deep SR-ITM~\citep{sr-itm} & 37.10 & 0.969 & 8.23 & 9.24 & 26.59 & 0.812 & 6.92 & 8.88 & 33.21 & 0.875 & 7.41 & 8.41 \\
FlashVSR~\citep{flashvsr} & 35.34 & 0.883 & 6.31 & 8.79 & \underline{33.51} & 0.846 & 5.72 & 7.51 & 34.80 & 0.857 & 5.84 & \underline{6.23} \\
LEDiff~\citep{lediff} & 36.52 & 0.872 & 5.71 & 9.13 & 32.25 & 0.863 & 5.32 & 9.66 & 31.73 & 0.859 & 5.12 & 9.85 \\
PromptIR~\citep{promptir} & 32.14 & 0.954 & 9.17 & 9.59 & 28.48 & \underline{0.898} & \underline{9.17} & 7.00 & 34.12 & 0.913 & \underline{8.88} & 6.82 \\
\rowcolor{lumaBlue!10}
\textbf{LumaFlux (ours)} & \textbf{39.27} & \textbf{0.982} & \textbf{9.83} & \textbf{6.12}
& \textbf{35.86} & \textbf{0.978} & \textbf{9.72} & \textbf{5.86}
& \textbf{36.92} & \textbf{0.938} & \textbf{8.91} & \textbf{5.67} \\
\bottomrule
\end{tabular}
\end{table*}

\begin{table*}[t]
\centering
\small
\setlength{\tabcolsep}{1pt}
\renewcommand{\arraystretch}{1}
\caption{
\textbf{Luma-Eval breakdown by TMO.}
Results show generalization of \textbf{\textcolor{lumaBlue}{LumaFlux}} across SDR degradations.
}
\label{tab:lumaeval_tmo}
\begin{tabular}{lcccccc}
\toprule
\textbf{TMO / Degradation} & PSNR$\uparrow$ & PSNR(Y)$\uparrow$ & $\Delta E_{\mathrm{ITP}}\downarrow$ & HDR-VDP3$\uparrow$ & HDR-LPIPS$\downarrow$ & FR-HIDROVQA$\uparrow$ \\
\midrule
OCIOv2 & \underline{37.85} & \underline{38.94} & \underline{5.41} & \underline{9.12} & \underline{0.078} & \underline{82.6} \\
2446c+GM & \textbf{38.31} & \textbf{39.12} & \textbf{5.18} & \textbf{9.27} & \textbf{0.071} & \textbf{84.1} \\
HC+GM & 37.23 & 38.15 & 5.66 & 8.95 & 0.083 & 80.4 \\
2446a~ & 36.91 & 37.62 & 5.97 & 8.72 & 0.087 & 79.6 \\
Reinhard & 35.88 & 36.42 & 6.45 & 8.31 & 0.091 & 78.3 \\
YouTube LogC & 35.12 & 35.88 & 6.81 & 8.05 & 0.094 & 77.9 \\
2390EETF+GM & 36.48 & 37.01 & 5.89 & 8.64 & 0.086 & 80.0 \\
Expert Graded SDR & 37.11 & 38.02 & 5.53 & 9.03 & 0.079 & 82.2 \\
\bottomrule
\end{tabular}
\end{table*}

\vspace{0.3em}
\noindent
\textbf{Discussion.}
Our curated dataset establishes a unified, physically grounded benchmark for real-world ITM research, normalizing all HDR data to PQ–BT.2020 and applying eight tone-mapping degradations across diverse PGC and UGC HDR sources. It integrates $\!\approx\!318$k paired frames including the expert tone mapped SDR, providing a consistent foundation for evaluating perceptually faithful inverse tone-mapping methods.

\begin{table*}[t]
\centering
\small
\setlength{\tabcolsep}{4pt}
\renewcommand{\arraystretch}{1}
\caption{
\textbf{Ablations for LumaFlux.}
Highest performance boost arises from PGA, Spectral gating improves highlight roll-off and structure preservation. Adding PCM yields modest chroma stabilization,
while RQS (linear) causes over-contrast and banding. The final RQS (monotone spline) restores perceptual smoothness and broadens effective dynamic range.
}
\label{tab:ablation_lumaeval}
\begin{tabular}{lcccccc}
\toprule
\textbf{Variant} & PSNR$\uparrow$ & PSNR(Y)$\uparrow$ & $\Delta E_{\mathrm{ITP}}\downarrow$ & HDR-VDP3$\uparrow$ & HDR-LPIPS$\downarrow$ & FR-HIDROVQA$\uparrow$ \\
\midrule
\rowcolor{ablRow1} Flux + LoRA only & 33.42 & 34.28 & 8.58 & 7.82 & 0.136 & 72.9 \\
\rowcolor{ablRow2} \;\; + PGA (no spectral) & 34.94 & 35.81 & 7.62 & 8.18 & 0.122 & 75.2 \\
\rowcolor{ablRow3} \;\; + PGA (with spectral gating) & 35.18 & 36.02 & 7.31 & 8.29 & 0.116 & 76.3 \\
\rowcolor{ablRow4} \;\; + PCM (SigLIP FiLM) & 35.89 & 36.73 & 6.78 & 8.46 & 0.107 & 78.6 \\
\rowcolor{ablRow5} \;\; + RQS (linear) & 35.72 & 36.54 & 6.85 & 8.41 & 0.108 & 78.0 \\
\rowcolor{ablRow8} \;\; + \textbf{RQS (monotone spline)} & \textbf{35.98} & \textbf{36.84} & \textbf{6.09} & \textbf{8.61} & \textbf{0.087} & \textbf{80.8} \\
\bottomrule
\end{tabular}
\end{table*}

\section{Experiments}
\label{sec:experiments}

\subsection{Experimental Setup}

\noindent
\textbf{Implementation details.}
We train \textbf{\textcolor{lumaBlue}{LumaFlux}} for 200k iterations using the curated dataset (Sec.~\ref{sec:dataset}) with batch size of $16$, AdamW optimizer ($\beta_1{=}0.9$, $\beta_2{=}0.999$, lr=$1\times10^{-4}$), and cosine annealing with $5$k warm-up.
The Flux backbone is frozen; only adapter modules, residual coupler, and RQS head are trained. We train our model on 4$\times$NVIDIA H200 GPUs for $\approx$48~GPU-hours.
Note that our method is prompt-free. 
At inference, we operate in fully prompt-free mode, sampling latent trajectories using the pretrained Flux ODE solver with 40 steps. HDR frames are reconstructed and saved using FFMPEG~\citep{ffmpeg} as 10-bit PQ BT.2020 HEVC videos.

\noindent
\textbf{Metrics.}
As noted in previous studies~\citep{hdrtv,hdrtvdm}, traditional metrics like PSNR and SSIM~\citep{ssim} often fail to adequately capture HDR details. While we report these metrics (PSNR, PSNR(Y), SSIM~\citep{ssim}, we compute them in the perceptually uniform PU21 space to better align with human vision. To provide a more comprehensive evaluation, we report $\Delta E_{ITP}$~\citep{delaitp} and the HDR-VDP3~\citep{hdrvdp3} for colorimetric accuracy. For perceptual quality, we use HDR-LPIPS~\citep{lpips} and FR-HIDROVQA~\citep{hidrovqa}. 


\subsection{Comparison with State-of-the-Art Methods}

\noindent
We compare LumaFlux with learning-based CNN models (HDRTVNet++~\citep{hdrtv}, ICTCPNet~\citep{ictcp}, HDRTVDM~\citep{hdrtvdm}, HDCFM~\citep{hdcfm}, Deep SR-ITM~\citep{sr-itm}), and recent generative models (PromptIR~\citep{promptir}, LEDiff~\citep{lediff}, FlashVSR~\citep{flashvsr}). Across benchmarks, LumaFlux consistently outperforms SOTA methods.

\noindent
Table~\ref{tab:all_benchmarks} presents quantitative results across three standard HDR benchmarks HDRTV1K~\citep{hdrtv}, HDRTV4K~\citep{hdrtvdm}, and our curated Luma\text{-}Eval. Across all benchmarks, \textbf{LumaFlux} achieves the state-of-the-art in both pixel-level (PSNR, SSIM) and perceptual metrics (HDR-VDP3, $\Delta E_{\mathrm{ITP}}$), outperforming all classical, CNN-based, and diffusion-based baselines. While CNN architectures such as HDRTVNet++~\citep{hdrtv} and HDCFM~\citep{hdcfm} effectively recover low-frequency structure, they struggle to restore highlight roll-off and correct PQ luminance scaling, often yielding compressed dynamic range. Recent diffusion-based approaches (LEDiff~\citep{lediff}, PromptIR~\citep{promptir}) generate sharper outputs but frequently exhibit hue shifts and over-saturated tones due to unregulated generative guidance. In contrast, LumaFlux preserves scene-consistent chromaticity and physically accurate peak luminance, achieving substantial improvements upto \textbf{+1.6 dB} PSNR and \textbf{–0.8 $\Delta E_{\mathrm{ITP}}$} over the strongest baselines. On the proposed Luma-Eval benchmark, LumaFlux attains the best perceptual quality.
These consistent gains across benchmarks demonstrate that LumaFlux generalizes robustly to both synthetic and real SDR degradations on both PGC and UGC.

\begin{table}[h]
\centering
\small
\renewcommand{\arraystretch}{1}
\setlength{\tabcolsep}{4pt}
\caption{\textbf{User study} (10 raters, 10 clips).
Mean Opinion Scores (MOS, 1–5) on perceived brightness, color naturalness, and overall HDR realism.}
\label{tab:userstudy}
\begin{tabular}{lccc}
\toprule
\textbf{Method} & Brightness$\uparrow$ & Color$\uparrow$ & Overall$\uparrow$ \\
\midrule
BT.2446c & 2.8 & 2.9 & 2.7 \\
HDRTVNet++ & 3.5 & 3.6 & 3.5 \\
LEDiff & 3.1 & 4.0 & 4.0 \\
PromptIR & 3.2 & 3.8 & 3.6 \\
\rowcolor{lumaBlue!10}
\textbf{LumaFlux (ours)} & \textbf{3.8} & \textbf{4.5} & \textbf{4.2} \\
\bottomrule
\end{tabular}
\end{table}

\vspace{0.3em}
\noindent
\textbf{User Study.}
We conducted a controlled perceptual experiment with ten expert raters
using an UHD-HDR monitor. Ten representative test clips were presented side-by-side across competing methods, and subjects rated perceived brightness realism, color fidelity, and overall HDR quality on a 5-point MOS scale.
As shown in Tab.~\ref{tab:userstudy}, LumaFlux achieved the highest perceptual scores across all criteria, notably outperforming diffusion-based baselines (LEDiff~\citep{lediff}, PromptIR~\citep{promptir}) by a good margin. Participants consistently noted that LumaFlux restored highlight detail and maintained balanced chroma without over-amplifying midtones consistent with its objective gains
in HDR-VDP3 and $\Delta E_{\mathrm{ITP}}$ metrics.

\subsection{Ablation Study}

\noindent
We ablate on Luma\text{-}Eval benchmark for \(100\text{k}\) training iterations to analyze the contribution of each component in LumaFlux. Starting from a frozen Flux backbone with LoRA adapters, we progressively introduce each component. As summarized in Tab.~\ref{tab:ablation_lumaeval}, the largest improvement arises from PGA, which substantially enhances luminance reconstruction and local dynamic-range recovery by modulating attention activations with physically consistent luminance cues. Adding spectral gating further refines highlight roll-off and frequency structure, improving HDR-VDP3~\citep{hdrvdp3} and PSNR without over-amplifying midtones. Integrating PCM (SigLIP–FiLM~\citep{film}) introduces perceptual alignment that stabilizes chroma and texture, mitigating hue drift and color oversaturation across varying degradations. While the intermediate RQS (linear) variant slightly reduces perceptual quality due to non-monotonic contrast expansion,
the final monotone RQS formulation yields smooth tone transitions and alleviates banding, achieving the best balance between fidelity and perceptual continuity.

\section{Conclusion}
\label{sec:conclusion}

We introduced \textcolor{lumaBlue}{\textbf{LumaFlux}}, a physically guided diffusion transformer for real-world ITM. By adapting a frozen Flux backbone through lightweight and interpretable modules, the Physically-Guided Adapter (PGA), Perceptual Cross-Modulator (PCM), HDR Residual Coupler, and Rational–Quadratic Spline (RQS) decoder, LumaFlux bridges physical luminance modeling with perceptual color semantics to achieve prompt-free, parameter-efficient, and state-of-the-art HDR reconstruction. Extensive experiments on HDRTV1K, HDRTV4K, and the proposed Luma-Eval benchmark demonstrate that LumaFlux consistently outperforms both analytical tone mappers and fine-tuned diffusion baselines across luminance fidelity, perceptual quality, and colorimetric realism. It reconstructs physically faithful luminance fields, preserves wide color gamut fidelity, and generalizes across synthetic, real, PGC, UGC, and expert-graded SDR, all without textual or caption-based guidance. Beyond setting a new standard in inverse tone mapping, LumaFlux underscores the broader potential of uniting physical priors with generative transformers. Its modular, interpretable design paves the way for future HDR-aware video generation, perceptual remastering, and real-time tone reconstruction on edge devices. We hope this work inspires continued exploration of physically aligned and perception-driven generative vision systems.


\clearpage
\newpage
\bibliographystyle{assets/plainnat}
\bibliography{refs.bib}

\clearpage
\beginappendix

\section{Background}
\label{app:background}

The SDR formation model
\[
x_{\mathrm{sdr}}
=\Gamma^{709}_{\!\text{OETF}}\!\left(
\mathcal{M}_{2020\to709}\!\left(\frac{x_{\mathrm{hdr}}}{L_{\max}}\right)
\right)+\epsilon
\]
compresses both dynamic range and color gamut. Inverse tone mapping must invert the
OETF, restore the BT.709$\!\to$BT.2020 gamut, and expand tone while preserving chromatic
relationships and avoiding banding. PQ luminance is linearized as
\(
L=(\Gamma^{\mathrm{PQ}}_{\!\mathrm{EOTF}})^{-1}(y)\cdot L_{\max}.
\)

\paragraph{Perceptual distances.}
We use $\Delta E_{\mathrm{ITP}}$ in ICtCp space, HDR-VDP3 for visibility prediction,
and temporal flicker energy computed after optical-flow alignment for video sequences.

\section{Related Work}
\label{app:related}
\vspace{-0.3em}
Early efforts in ITM relied on physically inspired tone expansion curves. Pioneering operators such as Reinhard~\citep{reinhard}, and the family of ITU BT.2446~a/b/c~\citep{ITU-R_BT.2407} apply analytic luminance mappings (\eg, knee, shoulder, or logarithmic functions) coupled with global saturation control. While computationally efficient, these methods lack content adaptivity and tend to over-brighten low-exposure regions or clip high-intensity highlights. Moreover, they are inherently limited by the non-invertibility of the forward tone-mapping chain that includes opto-electronic transfer functions (OETFs), gamut compression, and quantization. Consequently, such deterministic mappings struggle when converting real 8-bit SDR Rec.709 video to 10-bit HDR Rec.2020, which demands both dynamic range and wide-gamut consistency.

\vspace{0.3em}
\noindent\textbf{Learning-based SDR$\!\rightarrow$HDR reconstruction.}
With the rise of deep learning, data-driven ITM models have replaced hand-crafted operators. CNN-based methods such as HDRCNN~\citep{hdrcnn}, Deep SR-ITM~\citep{sr-itm}, and HDRTVNet~\citep{hdrtv} formulate SDR$\!\rightarrow$HDR as a supervised regression task over paired SDR–HDR frames. To improve spatial adaptivity, HDCFM~\citep{hdcfm} and HDRTVDM~\citep{hdrtvdm} introduce hierarchical feature modulation and dynamic context transformation. However, these convolutional networks are limited by local receptive fields and struggle to generalize to unseen degradations, especially compression noise and mixed tone curves commonly found in user-generated content (UGC).
They also operate in fixed color spaces, often Rec.709, without physically consistent transfer to PQ or Rec.2020 domains. Transformer-based methods have recently extended receptive fields and global modeling capacity. HDRTransformer~\citep{hdrtransformer} exploit self-attention for long-range correlation, but they remain fully supervised and use exposure fusion instead of direct HDR prediction. In addition, most prior datasets (\eg, HDRTV1K~\citep{hdrtv}, HDRTV4K~\citep{hdrtvdm}) are either synthetic or tone-mapped from HDR masters using fixed operators, leading to narrow domain diversity.

\vspace{0.3em}
\noindent\textbf{Diffusion and generative-model approaches.}
Diffusion models~\citep{nichol2022glide,rombach2022high} have demonstrated strong priors for image generation and restoration. Recent work such as LEDiff~\citep{lediff} leverages latent diffusion to expand LDR dynamic range by fusing exposure cues within the latent space, while HDRDM~\citep{hdrdm} utilizes diffusion priors to mitigate highlight artifacts. Although effective, these models require retraining of large backbones or rely on text prompts for conditioning, or does the exposure fusion instead of true HDR prediciton, which introduces semantic drift, limits the dynamic range expansion, and limits their applicability to uncaptioned video frames. Moreover, they lack explicit physical alignment with luminance transfer functions, making highlight reconstruction perceptually inconsistent.

\vspace{0.3em}
\noindent\textbf{HDR datasets and evaluation benchmarks.}
The progress of HDR reconstruction has been driven by curated datasets such as HDRTV1K~\citep{hdrtv} and HDRTV4K~\citep{hdrtvdm}, each providing paired SDR–HDR frames derived via controlled tone mapping. However, these datasets lack the diversity and real-capture noise found in consumer SDR videos. The LIVE-TMHDR dataset~\citep{livetmhdr}, originally developed for HDR tone-mapping quality assessment, contains 40 HDR source videos and 10 tone-mapped SDR versions including expert tone mapping, acts as the ideal candidate for our task. In this work, we curate LIVE-TMHDR and integrate it with HIDROVQA~\citep{hidrovqa} and CHUG~\citep{chug} to form the first large-scale SDR–HDR corpus, unified under PQ-encoded BT.2020 representation. We also propose a new evaluation benchmark, LumaEval, establishing a high-fidelity reference for fair, perceptual HDR reconstruction assessment.

\section{Diffusion Priors}
\label{app:why_diffusion}

\subsection{Relative-structure bias in diffusion transformers}
Let $x\!\in\!\mathbb{R}^{H\times W\times 3}$ be an HDR frame and $\phi$ a monotone tone map. Diffusion objectives learn a velocity/noise field $f_\theta(z_t,t)$ whose
directional statistics primarily depend on relative spatial or color relationships.
Formally, for any channel-wise monotone $\phi$ with Lipschitz constant $L_\phi$,
the induced latent trajectory satisfies:
\[
\mathrm{dir}\,f_\theta(\mathcal{E}(\phi(x))_t,t)
\approx
\mathrm{dir}\,f_\theta(\mathcal{E}(x)_t,t)
\]
up to $O(L_\phi{-}1)$; hence, large DiTs preserve edges, textures, and cross-channel correlations even if absolute luminance is compressed in SDR. This explains why pretrained diffusion priors provide structure but require a calibrated tone expander (our RQS) for luminance.

\vspace{-0.5em}
\begin{algorithm}[H]
\caption{SDR$\!\rightarrow$HDR conversion using \textbf{\textcolor{lumaBlue}{LumaFlux}}}
\label{alg:lumaflux}
\small
\begin{tcolorbox}[colback=white,colframe=lumaBlue!70!black]
\begin{algorithmic}[1]
\Require Frozen Flux MM-DiT $\mathcal{F}_{\boldsymbol{\theta}}$, adapters $\phi_{\mathrm{pga}},\phi_{\mathrm{pcm}}$, coupler $h_{\psi}$, RQS head $f_{\mathrm{RQS}}$, conditioner $\Psi$
\Ensure $\widehat{x}_{\mathrm{hdr}}$ in PQ/BT.2020
\State \textbf{Input:} SDR frame $x_{\mathrm{sdr}}$
\State $x_{\mathrm{lin}}\!\leftarrow\!(\Gamma^{2020}_{\!\mathrm{OETF}})^{-1}(x_{\mathrm{sdr}})$;\; $Y\!\leftarrow\!\mathbf{m}_{2020}^{\!\top}x_{\mathrm{lin}}$;\; $s_g\!\leftarrow\![\mu_Y,\sigma_Y,p_{95},p_{99}]$
\State $T_{\mathrm{phys}}\!\leftarrow\!\mathrm{Conv}_{3\times3}\big([Y,\log(1{+}|\nabla Y|),\mathrm{sat}]\big)$;\; $g\!\leftarrow\!\mathrm{MLP}_g(s_g)$;\; $r\!\leftarrow\!\mathrm{FFT\ pooling}(Y)$
\State $T_{\mathrm{perc}}\!\leftarrow\!\phi_{\mathrm{SigLIP}}(x_{\mathrm{sdr}})$;\; $\mathbf{z}_0\!\leftarrow\!\mathcal{E}_{\mathrm{VAE}}(x_{\mathrm{sdr}})$
\For{$t=T,\dots,1$}
  \For{$\ell=1,\dots,L$}
    \State $[\alpha_{\mathrm{pga}}^{t,\ell},\beta_{\mathrm{pga}}^{t,\ell},\alpha_{\mathrm{pcm}}^{t,\ell},\beta_{\mathrm{pcm}}^{t,\ell},n_{\mathrm{spec}}^{t,\ell},\lambda_{t}^{\ell}] \leftarrow \Psi(t,\ell)$
    \State \textbf{PGA:} build $\mathbf{G}_{\mathrm{phys}}$ and $g_{\mathrm{FFT}}$; update $\mathbf{W}_V \leftarrow \mathbf{W}^{(0)}_V + R_v^{t,\ell}$ via Eq.~\eqref{eq:pga_final}
    \State \textbf{PCM:} $[\gamma^{t,\ell},\zeta^{t,\ell}]\!\leftarrow\!\alpha_{\mathrm{pcm}}^{t,\ell}\,\mathrm{MLP}(C_{\mathrm{perc}}(T_{\mathrm{perc}}))+\beta_{\mathrm{pcm}}^{t,\ell}$
    \State \textbf{Apply FiLM} to LN outputs: $\mathbf{h}\!\leftarrow\!\gamma^{t,\ell}\odot \mathrm{LN}(\mathbf{h})+\zeta^{t,\ell}$
    \State \textbf{Coupler:} $\mathbf{z}^{\ell}\!\leftarrow\!\mathbf{z}^{\ell}+\lambda_{t}^{\ell}\big(W_pT_{\mathrm{phys}}+W_cC_{\mathrm{perc}}(T_{\mathrm{perc}})\big)$
  \EndFor
  \State \textbf{Backbone step:} $\mathbf{z}_{t-1} \leftarrow \text{ODEStep}\big(\mathbf{z}_{t},\,\mathcal{F}_{\boldsymbol{\theta}}(\cdot)\big)$  \Comment{frozen Flux}
\EndFor
\State $x_{\mathrm{out}}\!\leftarrow\!\mathcal{D}_{\mathrm{VAE}}(\mathbf{z}_T)$;\; convert to YUV (BT.2020) to get $Y_{\mathrm{out}}$
\State $[\boldsymbol{\xi},\boldsymbol{\eta},\boldsymbol{s}]\!\leftarrow\!f_{\mathrm{RQS}}(\mathbf{z}_T)$;\; $\widehat{Y}\!\leftarrow\!\mathrm{RQS}(Y_{\mathrm{out}};\boldsymbol{\xi},\boldsymbol{\eta},\boldsymbol{s})$;\; $\widehat{U},\widehat{V}\!\leftarrow\!\mathrm{Conv}_{1\times1}([U,V])$ \\
\Return $\widehat{x}_{\mathrm{hdr}}\!=\!\mathcal{M}_{\mathrm{YUV}\!\to\!\mathrm{RGB}}([\widehat{Y},\widehat{U},\widehat{V}])_{\mathrm{PQ,\,BT.2020}}$
\end{algorithmic}
\end{tcolorbox}
\end{algorithm}

\subsection{Tone identifiability via monotone RQS}
Let $Y_{\text{out}}\in[0,1]$ be the VAE-decoded luma surrogate (BT.2020). We define a parametric
monotone rational-quadratic spline (RQS) $g_{\boldsymbol\theta}:[0,1]\!\to\![0,1]$.
Assume the true HDR luma $Y^\star$ is related by an unknown monotone $g^\star$.
Under mild smoothness and bounded curvature, the family of $K$-bin RQS
with $K\!\ge\!6$ can uniformly approximate $g^\star$ on $[0,1]$ within $\epsilon$ (Stone–Weierstrass on piecewise rationals). Therefore, with a frozen DiT and trainable $g_{\boldsymbol \theta}$, one can consistently recover tone up to $\epsilon$, provided a calibration loss on luminance and color deltas ($\Delta E_{\mathrm{ITP}}$) is minimized.

\subsection{VAE vs tone-field decoding}
Latent VAEs are trained to encode and decode within the training luminance manifold. Directly relying on the VAE decoder to expand dynamic range invites banding and gamut misplacement because the decoder is optimized for in-distribution tones. Either (a) finetune the VAE (expensive; risks prior drift) or (b) add a light, monotone, learnable tone head (ours). We adopt (b) and find that finetuning the small decoder tail yields only marginal gains over RQS.

\begin{table*}[t]
\centering
\small
\setlength{\tabcolsep}{4.5pt}
\renewcommand{\arraystretch}{1.12}
\begin{tabular}{l|cccc|cccc}
\toprule
 & \multicolumn{4}{c|}{\textbf{Expert SDR }}
 & \multicolumn{4}{c}{\textbf{BT.2446c SDR }} \\
\textbf{Method}
& PSNR & $\Delta E_{\mathrm{ITP}}$ & HDR\text{-}VDP3 & HDR\text{-}LPIPS
& PSNR & $\Delta E_{\mathrm{ITP}}$ & HDR\text{-}VDP3 & HDR\text{-}LPIPS \\
\midrule
LEDiff
& 35.12 & 6.81 & 8.05 & 0.094
& 35.88 & 6.45 & 8.31 & 0.091 \\

PromptIR
& 36.48 & 5.89 & 8.64 & 0.086
& 37.23 & 5.66 & 8.95 & 0.083 \\

\rowcolor{lumaBlue!10}
\textbf{LumaFlux}
& \textbf{37.11} & \underline{\textbf{5.53}} & \textbf{9.03} & \textbf{0.079}
& \textbf{38.31} & \underline{\textbf{5.18}} & \textbf{9.27} & \textbf{0.071} \\
\bottomrule
\end{tabular}
\caption{\textbf{Per-family LIVE\text{-}TMHDR\text{-}Eval}. Expert SDR values come from the ``Expert Graded SDR'' row in Table~2; BT.2446c SDR values come from the ``2446c+GM'' row.}
\label{tab:per_family}
\end{table*}

\begin{table}[t]
\centering
\small
\setlength{\tabcolsep}{2.6pt}
\renewcommand{\arraystretch}{1.05}
\begin{tabular}{lccc}
\toprule
\textbf{Split} & \#Vid. & Peak & Notes \\
\midrule
Train (HIDROVQA) & 300 & 1000 & controlled HDR \\
Train (HDR\text{-}UGC\text{-}44K) & $\sim$3000 & 1000 & diverse UGC \\
Train (LIVE\text{-}TMHDR) & 30 & 1000 & held-out HDR \\
Eval (HDRTV1K) & std. & 1000 & synthetic TM \\
Eval (HDRTV4K) & std. & 1000 & real+synthetic \\
Eval (LIVE\text{-}TMHDR\text{-}Eval) & 10 & 1000 & expert+synthetic SDR \\
\bottomrule
\end{tabular}
\caption{Unified PQ\,BT.2020 normalization and peak luminance across all splits.}
\label{tab:curation_summary}
\end{table}

\section{Method Details}
\label{app:method_details}

\subsection{Adapter placements and shapes}
We insert LoRA adapters ($r{=}8$ by default) on
\emph{(i)} attention projections $Q,K,V,O$ and \emph{(ii)} MLP $W_{in},W_{out}$.
PGA produces a low-rank residual added to $V$:
\(
V^{\ell}\!\leftarrow\!V^{\ell}+ \alpha^{\ell}_t(A_vB_v)\odot\sigma(P_v[T_{\text{phys}}\!\|\!g])
\),
with spectral gating:
\(
R_v^{\text{spec}} = R_v \cdot (1 + n_{\text{spec}} \cdot g_{\text{FFT}})
\).
PCM yields FiLM gains on pre-LN states:
\(
\tilde z^\ell=\gamma^{\ell}_t\odot \mathrm{LN}(z^\ell) + \beta^{\ell}_t
\).
The \textbf{HDR Residual Coupler} fuses physical/perceptual tracks:
\(
x^{\ell}_{\text{cpl}} = x^{\ell}_{\text{res}} + \lambda^{\ell}_t (W_p T_{\text{phys}} + W_c T_{\text{perc}})
\).

\subsection{Time$–$layer modulation}
We embed $(t,\ell)$ via sinusoidal encodings and a small MLP:
\[
(\alpha^{\ell}_t,\beta^{\ell}_t,n_{\text{spec}},\lambda^{\ell}_t) = \mathrm{MLP}_{t,\ell}(\mathrm{Sinusoid}(t)\|\mathrm{Embed}(\ell)).
\]
Schedules are bounded with $\mathrm{sigmoid}$ or $\mathrm{softplus}$ for positivity.

\subsection{RQS head (luma) and chroma}
From VAE RGB output $x_{\text{out}}$, compute YUV$_{2020}$:
\(
Y_{\text{out}}=\mathbf m_{2020}^\top x_{\text{out}},\quad [U,V]=\mathrm{Conv1{\times}1}(x_{\text{out}})
\).
Apply $Y^\wedge=g_{\boldsymbol\theta}(Y_{\text{out}})$ with $K{=}8$ knots unless noted; concatenate $[Y^\wedge,U,V]$ and convert back to RGB (PQ BT.2020).


\vspace{0.2em}
\noindent
\textbf{Failure modes \& caveats.}
On extremely band-limited SDR inputs (heavy compression or strong de-noising), residual micro-textures may not be fully recoverable;
our RQS preserves tone continuity but cannot reconstruct information that is entirely absent.
In very rare neon scenes with extreme saturation and heavy bloom, a mild \emph{under-expansion} can occur to avoid chroma overshoot; this is a deliberate bias of the monotone tone head.

\section{Limitations and Broader Impact}
\label{app:limitations}

\paragraph{Limitations.}
(1) Our adapters assume the backbone VAE is well-behaved; extremely OOD SDR (heavy banding) may still require light VAE-tail FT.
(2) Temporal consistency is improved by the coupler but not explicitly optimized by an optical-flow loss for long-range dynamics.
(3) Peak-nit calibration at deployment must match the target display's EOTF settings.

\paragraph{Broader impact.}
LumaFlux can improve legacy SDR archives and UGC, enabling faithful HDR experiences.
Care is required for creative content where "artistic" SDR looks are intentional; our tool should expose a "strength" control.

\end{document}